\documentclass[twoside,11pt]{article}

\usepackage{jmlr2e}

\usepackage{amsthm,amssymb}
\usepackage{amsmath}
\usepackage{algorithm2e}
\usepackage{graphicx}
\usepackage{todonotes}
\usepackage{subcaption}
\usepackage{multirow}
\usepackage{tikz}
\usepackage{arydshln}
\usepackage{lineno}

\usepackage{lastpage}

\ShortHeadings{Sparse Adaptive Bottleneck Centroid-Encoder}{Ghosh and Kirby}
\firstpageno{1}

\begin{document}

\title{Feature Selection using Sparse Adaptive Bottleneck Centroid-Encoder}

\author{\name Tomojit Ghosh \email Tomojit.Ghosh@colostate.edu \\
       \addr Department of Mathematics\\
       Colorado State University\\
       Fort Collins, CO 80523 ,USA
       \AND
       \name Michael Kirby \email Kirby@math.colostate.edu \\
       \addr Department of Mathematics\\
       Colorado State University\\
       Fort Collins, CO 80523, USA}
\editor{TBD}
\maketitle

\begin{abstract}
We introduce a novel nonlinear model, Sparse Adaptive Bottleneck Centroid-Encoder (SABCE), for determining the features that discriminate between two or more classes. The algorithm aims to extract discriminatory features in groups while reconstructing the class centroids in the ambient space and simultaneously use additional penalty terms in the bottleneck layer to decrease within-class scatter and increase the separation of different class centroids. The model has a sparsity-promoting layer (SPL) with a one-to-one connection to the input layer. Along with the primary objective, we minimize the $l_{2,1}$-norm of the sparse layer, which filters out unnecessary features from input data. During training, we update class centroids by taking the Hadamard product of the centroids and weights of the sparse layer, thus ignoring the irrelevant features from the target. Therefore the proposed method learns to reconstruct the critical components of class centroids rather than the whole centroids. The algorithm is applied to various real-world data sets, including high-dimensional biological, image, speech, and accelerometer sensor data. We compared our method to different state-of-the-art feature selection techniques, including supervised Concrete Autoencoders (SCAE), Feature Selection Networks (FsNet), Stochastic Gates (STG), and LassoNet. We empirically showed that SABCE features often produced better classification accuracy than other methods on the sequester test sets, setting new state-of-the-art results.
\end{abstract}

\section{Introduction}
\label{intro}
Technological advancement has made high-dimensional data readily available. For example, in bioinformatics, the researchers seek to understand the gene expression level with microarray or next-generation sequencing techniques where each point consists of over 50,000 measurements \citep{Pease5022,shalon1996dna,metzker2010sequencing,reuter2015high}. 
The abundance of features demands the development of feature selection algorithms to improve a Machine Learning task, e.g., classification. Another important aspect of feature selection is knowledge discovery from data. Which biomarkers are important to characterize a biological process, e.g., the immune response to infection by respiratory viruses such as influenza \citep{o2013iterative}? Additional benefits of feature selection include improved visualization and understanding of data, reducing storage requirements, and faster algorithm training times.

Feature selection can be accomplished in various ways that can be broadly categorized as filter, wrapper, and embedded methods. In a filter method, each variable is ordered based on a score. After that, a threshold is used to select the relevant features \citep{lazar2012survey}. Variables are usually ranked using correlation \citep{guyon2003introduction,yu2003feature}, and mutual information \citep{vergara2014review,fleuret2004fast}. In contrast, a wrapper method uses a model and determines the importance of a feature or a group of features by the generalization performance of the predetermined model \citep{el2016review, hsu2002annigma}. Since evaluating every possible combination of features becomes an NP-hard problem, heuristics are used to find a subset of features. Wrapper methods are computationally intensive for larger data sets, in which case search techniques like Genetic Algorithm (GA) \citep{goldberg1988genetic} or Particle Swarm Optimization (PSO) \citep{kennedy1995particle} are used. In embedded methods, feature selection criteria are incorporated within the model, i.e., the variables are picked during the training process \citep{lal2006embedded}. Iterative Feature Removal (IFR) uses the ratio of absolute weights of a Sparse SVM model as a criterion to extract features from the high dimensional biological data set \citep{o2013iterative}.

Mathematically feature selection problem can be posed as an optimization problem on $\ell_0$-norm, i.e., how many predictors are required for a machine learning task. As the minimization of $\ell_0$ is intractable (non-convex and non-differentiable), $\ell_1$-norm is used instead, which is a convex proxy of $\ell_0$ \citep{tibshirani1996regression}. Although the $\ell_1$ has been used in the feature selection task in linear \citep{fonti2017feature,muthukrishnan2016lasso,kim2004gradient,o2013iterative,chepushtanova2014band} as well as in nonlinear regime \citep{li2016deep,scardapane2017group,li2020efficacy}, it has some disadvantages as well. For example, when multi-collinearity exists (i.e., two or more independent features have a high correlation with one another) in data, $\ell_1$ selects one of them and discards the rest, degrading the rest prediction performance \citep{zou2005regularization}. Although the problem can be overcome using iterative feature removal scheme as proposed by \citep{o2013iterative}. It has been reported that minimizing Lasso doesn't satisfy the Oracle property \citep{zou2006adaptive}. ElasticNet \citep{zou2005regularization}, on the other hand, overcomes some limitations of Lasso by combining $\ell_2$ norm with $\ell_1$-norm.

This paper proposes a new embedded variable selection approach called Sparse Adaptive Bottleneck Centroid-Encoder (SABCE) to extract features when class labels are available. Our method modifies the Centroid-Encoder model \citep{GHOSH201826,ghosh2022supervised} by incorporating two penalty terms in the bottleneck layer to increase class separation and localization. SABCE applies a $\ell_{2,1}$ penalty to a sparsity-promoting layer between the input and the first hidden layer while reconstructing the class centroids. One key attribute of SABCE is the adaptive centroids update during training, distinguishing it from Centroid-Encoder, which has a fixed class centroid. We evaluate the proposed model on diverse data sets and show that the features produce better generalization than other state-of-the-art techniques.

\section{Sparse Adaptive Bottleneck Centroid-Encoder (SABCE)}
\label{SABCE}

Consider a data set $X=\{x_i\}_{i=1}^{N}$ with $N$ samples and $M$ classes where $x_i \in \mathbb{R}^d$. The classes denoted $C_j, j = 1, \dots, M$ where the indices of the data associated with class $C_j$ are denoted $I_j$. We define centroid of each class as $c_j=\frac{1}{|C_j|}\sum_{i \in I_j} x_i$ where $|C_j|$ is the cardinality of  class $C_j$.

\subsection{Bottleneck Centroid-Encoder (BCE)}

Given the setup mentioned above, we define Bottleneck Centroid-Encoder, which is the starting point of our proposed algorithm. The objective function of BCE is given below:
\begin{equation}
\begin{aligned}
 \mathcal{L}_{bce}(\theta)=\frac{1}{2N}\sum^M_{j=1} \sum_{i \in I_j}(\|c_j-f(x_i; \theta)\|^2_2 + \mu_1\|g(c_j)-g(x_i)\|^2_2)
 + \mu_2\sum_{k < l}\frac{1}{1+\|g(c_k)-g(c_l))\|^2_2}
 \label{equation:ModifiedCECostFunction}
\end{aligned}
\end{equation}
The mapping $f$ is composed of a dimension-reducing mapping $g$ (encoder) followed by a dimension-increasing reconstruction mapping $h$ (decoder). The first term of the objective is minimizing the square of the distance between $f(x_i)$ and its class centroid $c_j$. Therefore the aim is to map the sample $x_i$ to its corresponding class centroid $c_j$, and the mapping function $f$ is known as Centroid-Encoder \citep{ghosh2022supervised}. The output of the encoder $g$ is used as a supervised visualization tool \citep{ghosh2022supervised,GHOSH201826}. Centroid-Encoder calculates its cost on the output layer; if the centroids of multiple classes are close in ambient space, the corresponding samples will land close in the reduced space, increasing the error rate. To remedy the situation, we add two more terms to the bottleneck layer, i.e., at the output of the encoder $g$, which we call Bottleneck Centroid-Encoder (BCE). The term $\|g(c_j)-g(x_i)\|^2_2$ will further pull a sample $x_i$ towards it centroid which will improve the class localization in reduced space. 
Further, to avoid the overlap of classes in the latent
bottleneck space, we introduce a term that serves to
to repel centroids there. 
We achieve this by maximizing the distances (equivalently the square of the $\ell_2$-norm) between all class-pairs  of latent centroids. We introduced the third term to fulfill the purpose. Note, as the original optimization is a minimization problem, we choose to minimize $\sum_{k < l}\frac{1}{1+\|g(c_k)-g(c_l)\|^2_2}$ which will ultimately increase the distance between the latent centroids of class $k$ and $l$. We added $1$ in the denominator for numerical stability. The hyper parameters $\mu_1$ and $\mu_2$ will control the class localization and separation in the embedded space. We use a validation set to determine their values.

\subsection{Sparse Adaptive Bottleneck Centroid-Encoder for Robust Feature Selection}

\begin{figure}[h!]
	\centering
	\includegraphics[width=13.75cm,height=5.75cm]{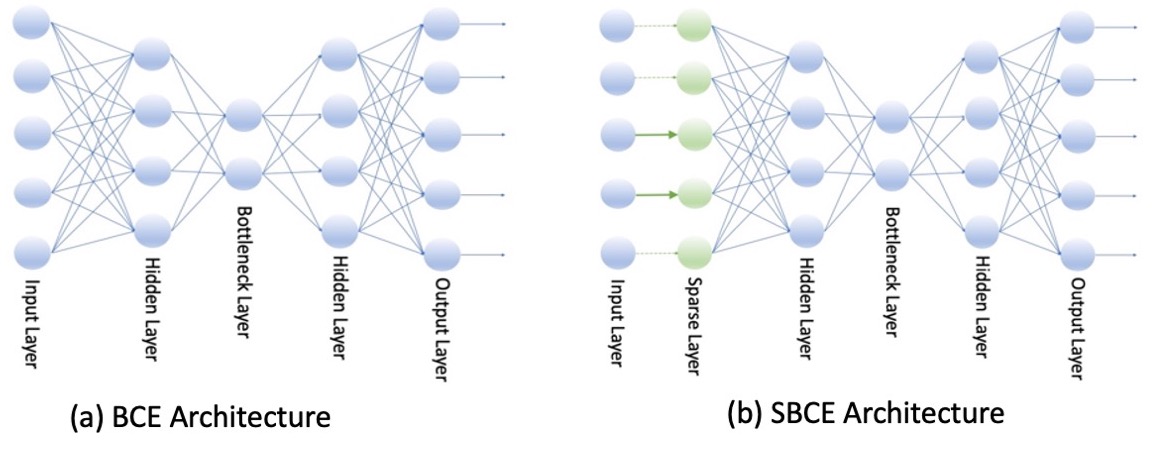}
	\caption{The architecture of Bottleneck Centroid-Encoder and Sparse Bottleneck Centroid-Encoder. Notice that Sparse Bottleneck Centroid-Encoder employs a sparse layer between the input and the first hidden layer to promote feature sparsity using $\ell_{2,1}$ norm.}
	\label{fig:SBCE_arch}
\end{figure}

The Sparse Bottleneck Centroid-Encoder (SBCE) is a modification to the BCE architecture as shown in Figure \ref{fig:SBCE_arch}. The input layer is connected to the first hidden layer via the sparsity promoting layer (SPL). Each node of the input layer has a weighted one-to-one connection to each node of the SPL. The number of nodes in these two layer are the same. The nodes in SPL don't have any bias or non-linearity. The SPL is fully connected to the first hidden layer, therefore the weighted input from the SPL will be passed to the hidden layer in the same way that of a standard feed forward network. During training, an $\ell_{2,1}$ penalty, which is also known as Elastic Net \citep{zou2005regularization}, will be applied to the weights connecting the input layer and SPL layer. The sparsity promoting $\ell_{2,1}$ penalty will drive most of the weights to near zero and the corresponding input nodes/features can be discarded. Therefore, the purpose of the SPL is to select important features from the original input. Note we only apply the $\ell_{2,1}$ penalty to the parameters of the SPL.

Denote $\theta_{spl}$ to be the parameters (weights) of the SPL and $\theta$ to be the parameters of the rest of the network. The cost function of sparse bottleneck centroid-encoder is given by
\begin{equation}
\begin{aligned}
 \mathcal{L}_{sbce}(\theta,\theta_{spl})=\frac{1}{2N}\sum^M_{j=1} \sum_{i \in I_j}(\|c_j-f(x_i; \theta)\|^2_2 + \mu_1\|g(c_j)-g(x_i)\|^2_2) \\
 + \mu_2\sum_{k < l}\frac{1}{1+\|g(c_k)-g(c_l))\|^2_2} + \lambda_1 \|\theta_{spl}\|_{1} + \lambda_2 \|\theta_{spl}\|^2_{2}
 \label{equation:SBCECostFunction}
\end{aligned}
\end{equation}
where $\lambda_1,\lambda_2$ are the hyperparameter which control the sparsity. A larger value of $\lambda_1$ will promote higher sparsity resulting more near-zero weights in SPL. 

\subsubsection{Sparsification of the Centroids}
The targets of the SBCE are the class centroids which are pre-computed from data and labels. For high-dimensional datasets, the features are noisy, redundant, or irrelevant \citep{alelyani2018feature}; therefore, feature selection with fixed class centroids, computed on high-dimensional ambient space, may be impacted by the noise. We can remedy the situation by 
promoting sparsity in the class centroid during training. In this approach, we start with $c_j$'s computed in the ambient space; after that, we change $c_j$'s by multiplying it component-wise by $\theta_{spl}$, i.e., $[{c_j}]_{t} = [{c_j}] \odot [{{\theta}_{spl}}]_{t-1}$, where $t$ is the current epoch. As $\theta_{spl}$ sparsifies the input data by eliminating redundant and noisy features, updating the centroids as shown above will reduce noise from the targets, thus improving the discriminative power of selected features. We call this algorithm as Sparse Adaptive Bottleneck Centroid-Encoder (SABCE). In Equation \ref{equation:SBCECostFunction}, we use $[{c_j}]_{t}$ (instead of $c_j$) to calculate the cost at each iteration. The comparison in Table \ref{table:SBCE_vs_SABCE} shows the performance advantage of SABCE over SBCE. 

\begin{table}[ht!]
	\vspace{-4mm}
	\centering
	\begin{tabular} {|c|c|c|c|c|c|}	
		\hline
		\multirow{2}{*}{Models} & \multicolumn{5}{c|} {Data set} \\
		\cline{2-6}
		& \multicolumn{1}{c|} {Mice Protein} & \multicolumn{1}{c|} {MNIST} & \multicolumn{1}{c|} {FMNIST} &  \multicolumn{1}{c|} {GLIOMA} & \multicolumn{1}{c|} {Prostate\_GE} \\
		
		\hline
		SBCE & $95.8$ & $92.5$ & $85.0$ & $65.6$ & $88.2$ \\
		\hline
		SABCE & \textbf{99.8} & $\textbf{94.0}$ & $\textbf{85.4}$ & \textbf{74.2} & \textbf{90.2}\\
		\hline

	\end{tabular}	
 \vspace{1mm}
	\caption{Comparison between SBCE and SABCE on five benchmarking data sets using top 50 features. We use the same network architecture and hyper parameters in training. We follow the same experimental set up as in Section \ref{experiments}.}
	\label{table:SBCE_vs_SABCE}
\vspace{-0.5cm}
\end{table}

\textbf{Training Details:} We implemented SLCE in PyTorch \citep{paszke2017automatic} to run on GPUs on Tesla V100 GPU machines. We trained SABCE using Adam \citep{DBLP:journals/corr/KingmaB14} on the whole training set without minibatch. We pre-train our model for ten epochs and then include the sparse layer (SPL) with the weights initialized to 1. Then we train the model for another ten epochs to adjust the weights of the SPL. After that, we did an end-to-end training applying $\ell_{2,1}$-penalty on the SPL for 1000 epochs.
Like any neural network-based model, the hyperparameters of SABCE need to be tuned for optimum performance. Table \ref{table:SBCEHyperparameters} contains the list with the range of values we used in this paper. We used validation set to choose the optimal value.  Section \ref{reproducibility} of Appendix has information on reproducibility. We will provide the code with a dataset as supplementary material. 

\begin{table}[!ht]	
	\vspace{1mm}
	\centering
	\begin{tabular} {|c|c|}
		\hline 	
		Hyper parameter & Range of Values \\
		\hline
		\# Hidden Layers (L) & \{1, 2\} \\
		\hline
        \# Hidden Nodes (H) & \{50,100,200,250,500\} \\
        \hline
        Activation Function & Hyperbolic tangent (tanh) \\
        \hline
        $\mu_1,\mu_2$ & \{0.1, 0.2, 0.3, 0.4, 0.5, 0.6, 0.7\} \\
        \hline
        $\lambda_1,\lambda_2$ & \{0.01, 0.001, 0.0001, 0.0002, 0.0004, 0.0006, 0.0008\} \\
        \hline
	\end{tabular}
 \vspace{1mm}
	\caption{Hyperparameters for Sparse Adaptive Bottleneck Centroid-Encoder.}
	\label{table:SBCEHyperparameters}
\end{table}

\subsubsection{Feature Cut-off}
\label{feature_cut-off}
\begin{figure}[h!]
    \centering
    \includegraphics[width=11.0cm, height=4.5cm]{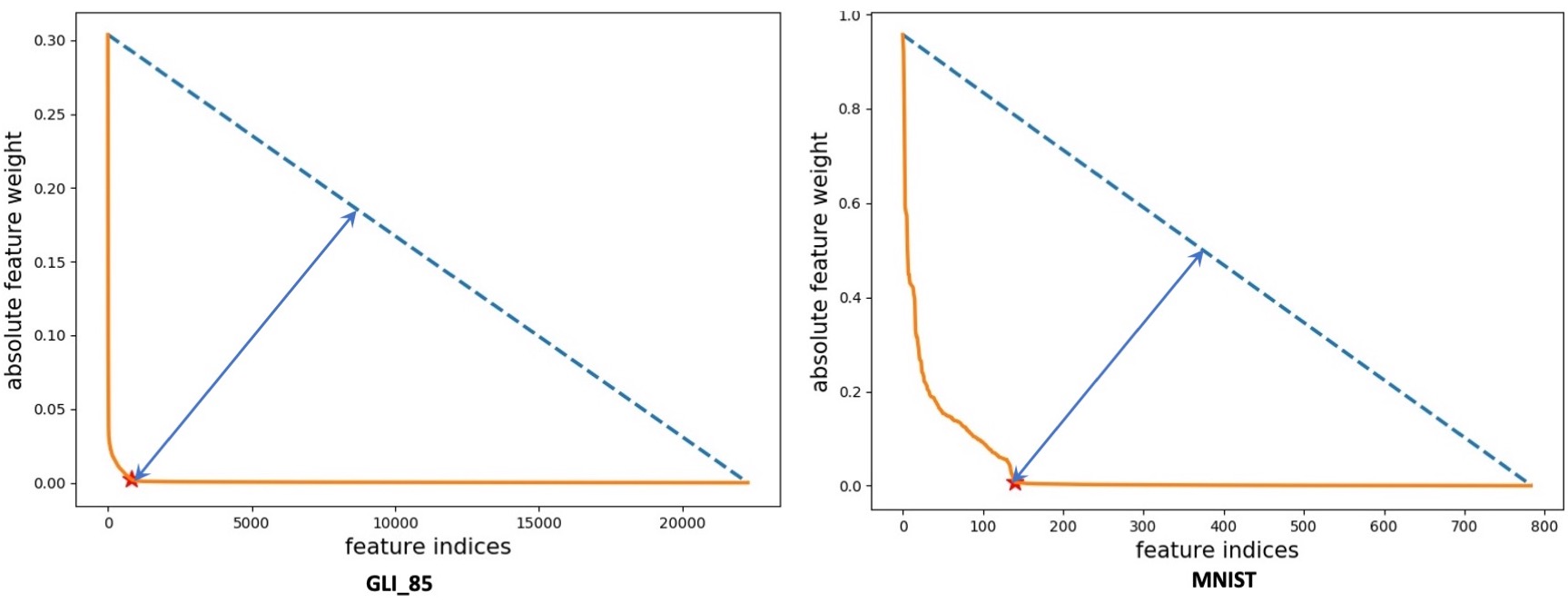}
    \vspace{1mm}
    \caption{Feature selection cut off geometry.}
    \label{fig:Feature_Cutoff_Elbow}
\end{figure}

As shown in Figure \ref{fig:Feature_Cutoff_Elbow},
the $\ell_{2,1}$-norm of the sparse layer (SPL) drives a lot of weights to near zero. Often hard thresholding or a ratio of two consecutive weights is used to pick the nonzero weight \citep{o2013iterative}.

In this article, we take a different approach to select the set of discriminatory features as shown in Figure \ref{fig:Feature_Cutoff_Elbow}. After training SBCE, we arrange the absolute value of the weights of the sparse layer in descending order forming a curve (the orange one). We then join the first and the last point of the curve with a straight line (the blue dotted line). We measure the distance of each point on the curve to the straight line. The point with the largest distance is the position (P) of the elbow. We pick all the features whose absolute weight is greater than that of P.  
 Figure \ref{fig:Feature_Cutoff_Elbow} demonstrates the approach on GLI\_85 and MNIST set. The red star indicates the position of point P (the elbow), and the absolute weight of features on the left of P is higher than on the right, selecting only 796 out of 22,283 features from GLI\_85 and 137 out of 784 MNIST pixels.

\subsection{Empirical Analysis of SBCE}
\label{empirical_analysis}
In this section we present a series of analyses of the proposed model to understand its behavior.

\begin{figure}[ht!]
    \centering
    \vspace{-0.10cm}
    \includegraphics[width=10.5cm, height=6.0cm]{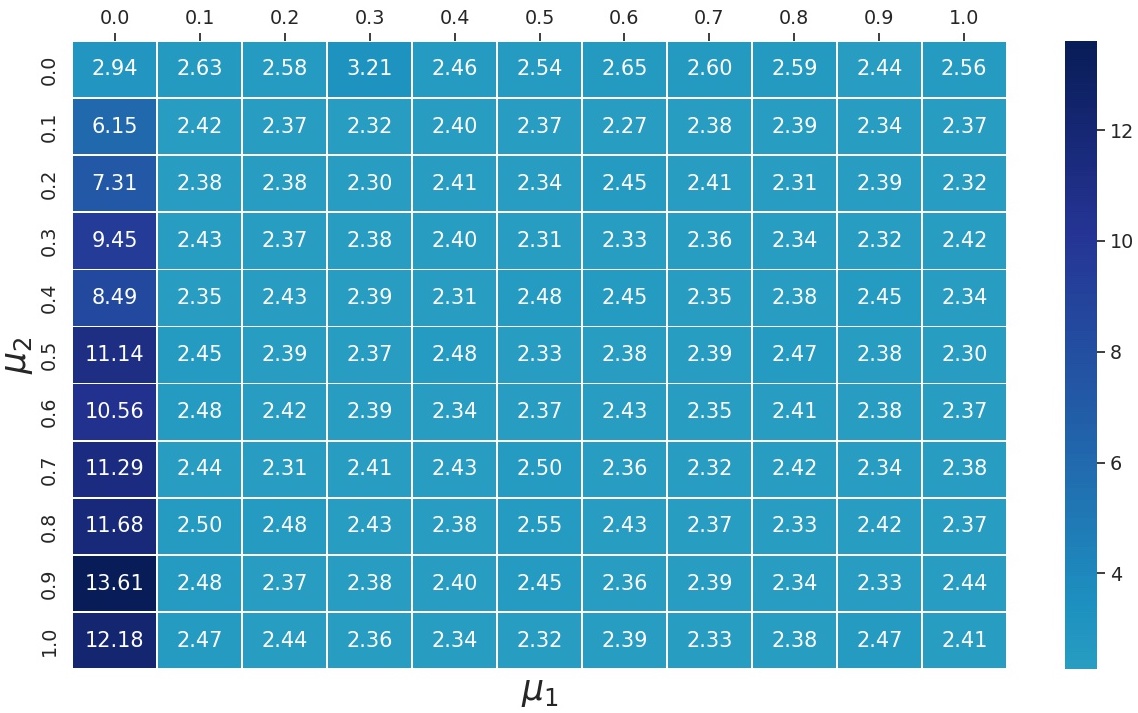}
    \vspace{-0.4cm}
    \caption{Analysis of error rate with changes to $\mu_1$ and $\mu_2$.}
    \label{fig:SBCE_mu1_mu2_analysis}
\end{figure}

\textbf{1. Analysis of Hyper-parameters $\mu_1$ and $\mu_2$: }
The hyper-parameters $\mu_1$ and $\mu_2$ control the class scatter and separation in bottleneck space. We ran an experiment on the MNIST digits to understand the effect of $\mu_1$ and $\mu_2$ on model's performance. We put aside $20\%$ of samples from each class as a validation set and the rest of the data set is used to train a the model for each combination of $\mu_1$ and $\mu_2$. The validation set is used to compute the error rate using a 5-NN classifier in the two-dimensional space. Figure \ref{fig:SBCE_mu1_mu2_analysis} shows the errors for different combinations of $\mu_1$ and $\mu_2$ in a heat map.
Observe that the error rate increases with $\mu_2$ when $\mu_1$ is zero. The behavior is not surprising as setting $\mu_1$ to zero nullifies the effect of the second term (see Equation \ref{equation:SBCECostFunction}), which would hold the samples tightly around their centroid in reduced space. The gradient from the first term will exert a pulling force to bind the samples around their centroids, but the gradient coming from the third term will dominate the gradient of the first term as $\mu_2$ increases. As an effect, the class-scatter increases in low dimensional space resulting in misclassifications. As soon as $\mu_1$ increases to 0.1, the error rate decreases significantly. After that, a higher value of $\mu_2$ doesn't change the results too much. The minimum validation error occurs for $\mu_1=0.6$ and $\mu_2=0.1$. The analysis reveals that $\mu_1$ is relatively more important than $\mu_2$.

\textbf{2. Analysis of Feature Sparsity:} 
\begin{figure}[ht!]
    \centering
    \vspace{-0.10cm}
    \includegraphics[width=10.5cm, height=6.0cm]{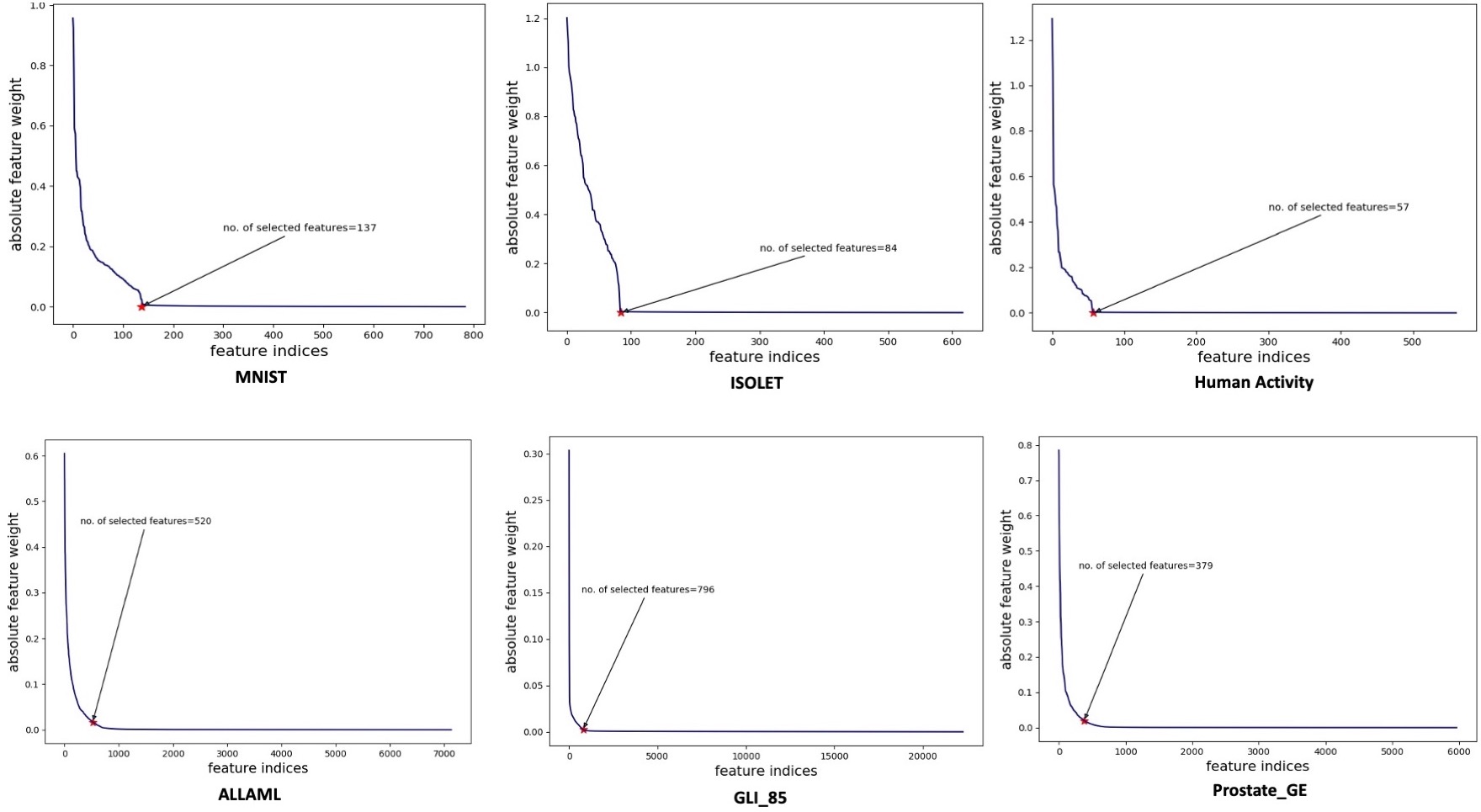}
    \vspace{-0.3cm}
    \caption{Sparsity analysis of SABCE on six data sets. MNIST, ISOLET, and Human Activity have more samples than features while ALLAML, GLI\_85, and Prostate\_GE have more features than data points; see Table \ref{table:dataDescription} for more details. In each case, we plotted the absolute value of weights of the sparse layer in descending order. In all the experiments, we set $\lambda_1,\lambda_2,$ to 0.001. These plots suggest the model promotes sparsity, driving most of the weights in the sparsity layer to near zero.}
    \label{fig:SBCE_sparsity_analysis}
\end{figure}
Here, we study sparsity promotion on six
representative data sets.
Three datasets contain more samples
than features: MNIST, ISOLET, and Human Activity. These data sets are also from different domains, namely image, speech, and accelerometer sensors. We also use three biological data sets: ALLAML, GLI\_85, and Prostate\_GE, where the sample size is significantly smaller when compared to the number of features (see Table \ref{table:dataDescription}). We fit our model on the training partition of each data set and then plot the absolute value of the weights of the sparse layer in descending order as shown in Figure \ref{fig:SBCE_sparsity_analysis}. As can be seen, the model promotes sparsity in each case, driving most of the weights in the sparse layer to near zero ($10^{-4}$ to $10^{-6}$). The rest of the features have significantly higher values, and our feature cut-off technique distinguishes them successfully, pointing out the number of selected features in each case.


\textbf{3. Effect of $\lambda_1$ and $\lambda_2$ on Feature Sparsity:} 
\begin{figure}[ht!]
    \centering
    \vspace{-0.10cm}
    \includegraphics[width=12.0cm, height=5.0cm]{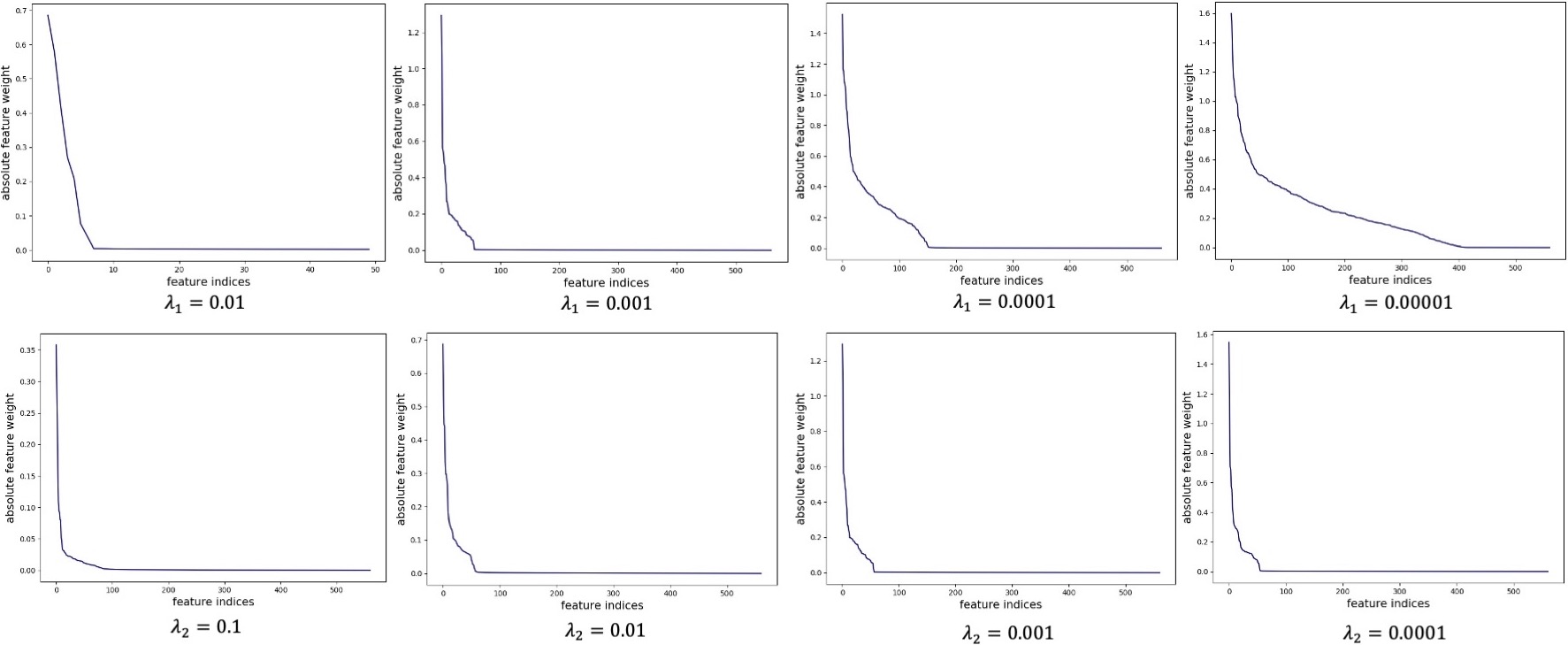}
    \vspace{-0.30cm}
    \caption{Effect of $\lambda_1$ and $\lambda_2$ on sparsity.}
    \label{fig:SBCE_L1_vs_L2_Sparsity}
\end{figure}
In Figure \ref{fig:SBCE_L1_vs_L2_Sparsity}, we show how hyperparameters $\lambda_1$ and $\lambda_2$ control sparsity on Human Activity data. We fix $\lambda_2$ to $0.001$ and run SABCE with different values of $\lambda_1$, which we show in the first row. Observe that the solution become less sparse with the decrease of $\lambda_1$. In the second row, we present a similar plot over different values of $\lambda_2$ while fixing $\lambda_1$ to $0.001$. The change of $\lambda_2$ doesn't contribute too much to the model's sparsity.

\begin{figure}[ht!]
    \centering
    \vspace{-0.1cm}
    \includegraphics[width=10.5cm, height=6.0cm]{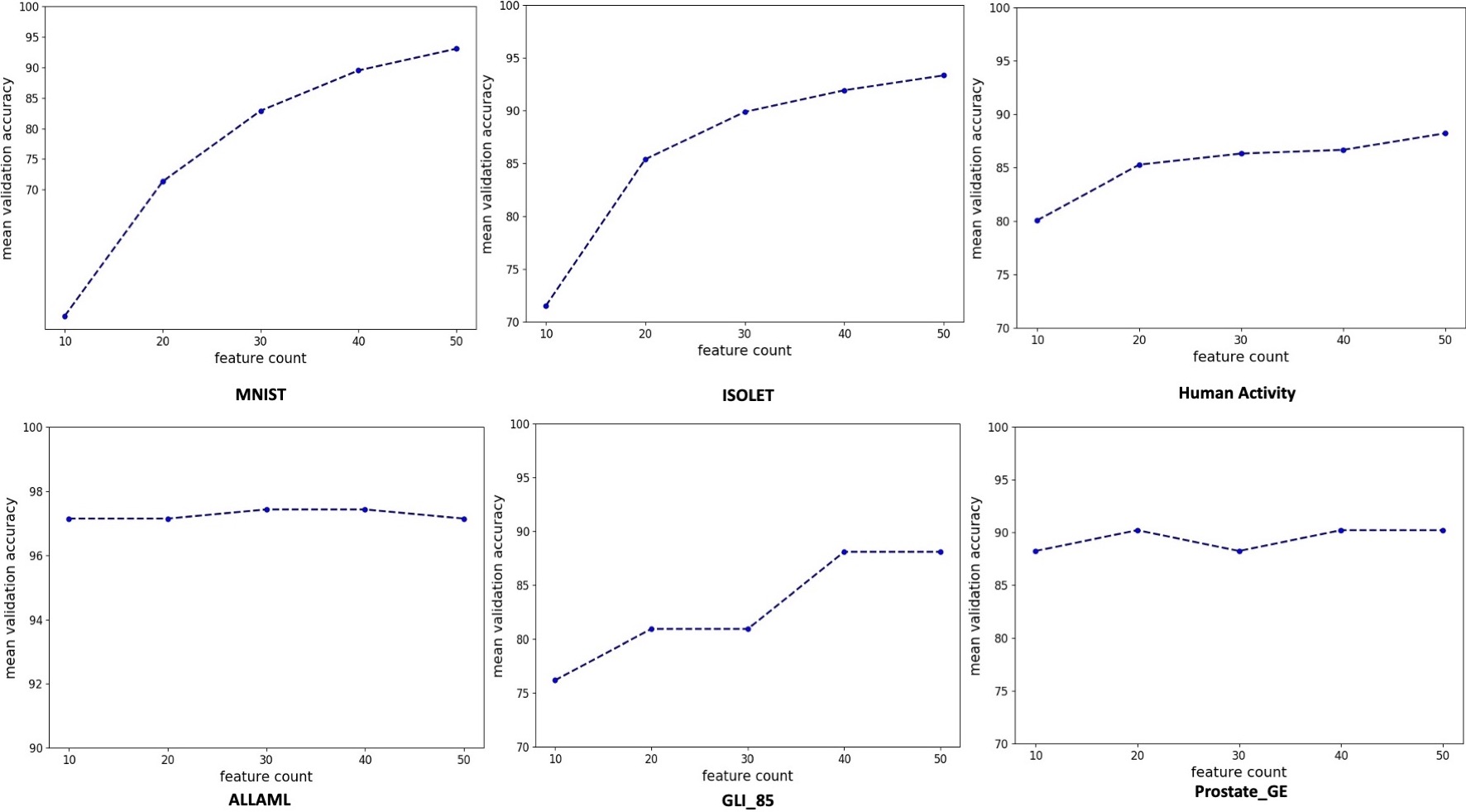}
    \vspace{-0.3cm}
    \caption{Accuracy on the validation set as a function of the number of features.}
    \label{fig:SBCE_Accuracy_vs_Feature_Cnt}
\end{figure}

\newpage
\textbf{4. Generalization Property of Selected Features:} To investigate the generalization aspect of the features, we restrict training and validation sets to the selected variables. We fit a one-hidden layer neural network on the training set to predict the class label of the validation samples. Figure \ref{fig:SBCE_Accuracy_vs_Feature_Cnt} shows the accuracy as a function of feature count. 

The plot suggests that the selected features can accurately predict unseen samples. We see that the accuracy increases with the number of features for MNIST, ISOLET Human Activity, and GLI\_85; in contrast, adding more features does not change the accuracy for ALLAML and Prostate\_GE significantly.

\textbf{5. Feature Selection Stability:} In this experiment, we shed light on the stability of the feature selection process; specifically, we want to compare and contrast the feature sets across several trials. To this end, We run our model five times on two high-dimensional biological data, ALLAML, Prostate\_GE, and one high sample size data Human Activity and then compare the feature sets. 
\begin{figure}[h!]
    \centering
    \includegraphics[width=13.0cm, height=4.0cm]{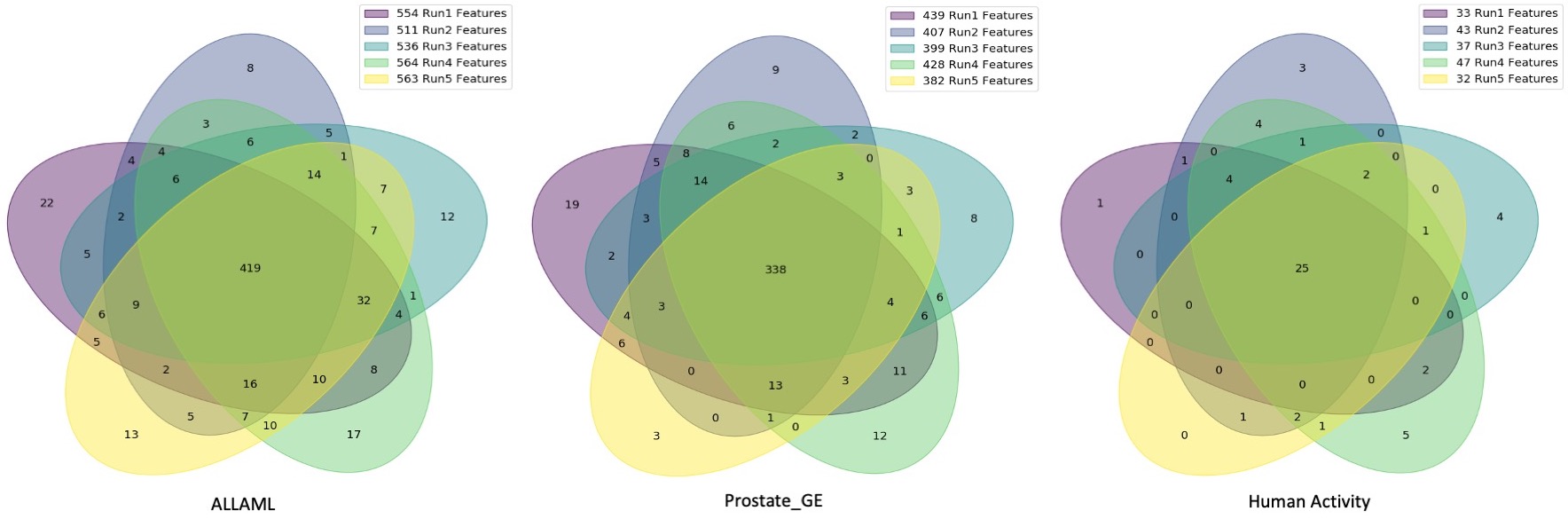}
    \caption{Venn diagram of five sets of features for each of the three high-dimensional data sets. }
    \label{fig:SABCE_Feature_Stability}
\end{figure}

Figure \ref{fig:SABCE_Feature_Stability} shows the results using Venn diagrams. Observe that the number of selected features over five runs are generally close to each other. In each case, there is a significant number of overlapping features. For each data set, we calculate the Jaccard index using the five feature sets to measure their similarity—the Jaccard index of ALLAML, Prostate\_GE, and Human Activity are 0.6254, 0.6828, and 0.6253, respectively. High Jaccard scores indicate that the feature sets have a lot of commonality over different runs.

\section{Experimental Results}
\label{experiments}
We present the comparative evaluation of our model on various data sets using several feature selection techniques. 

\subsection{Experimental Details}

\begin{table}[!ht]	
	\vspace{-4mm}
	\centering
	\begin{tabular} {|c|c|c|c|c|c|}	
		\hline	
		Dataset & No. Features & No. of Classes & No. of Samples & Domain \\
		\hline
		ALLAML & 7129 & 2 & 72 & Biology \\
		GLIOMA & 4434 & 4 & 50 & Biology \\
		SMK\_CAN & 19993 & 2 & 187 & Biology \\
		Prostate\_GE & 5966 & 2 & 102 & Biology \\
		GLI\_85 & 22283 & 2 & 85 & Biology \\
		CLL\_SUB & 11340 & 3 & 111 & Biology \\
		Mice Protein & 77 & 8 & 975 & Biology \\
		COIL20 & 1024 & 20 & 1440 & Image \\
		Isolet & 617 & 26 & 7797 & Speech \\
		Human Activity & 561 & 6 & 5744 & Accelerometer Sensor \\
		MNIST & 784 & 10 & 70000 & Image \\
		FMNIST & 784 & 10 & 70000 & Image \\
		\hline
	\end{tabular}	
	\caption{Descriptions of the data sets used for benchmarking experiments.}
	\label{table:dataDescription}
\end{table}

We used twelve data sets from a variety of domains (image, biology, speech, and sensor; see  
Table \ref{table:dataDescription}) and five neural network-based models to run three benchmarking experiments. To this end, we picked the published results from two papers \citep{lemhadri2021lassonet, singh2020fsnet} for benchmarking and we ran the Stochastic Gate \citep{yamada2020feature} using the code provided by authors. We followed the same experimental methodology described in \citep{lemhadri2021lassonet, singh2020fsnet} for an apples-to-apples comparison. This approach permitted a direct comparison of LassoNet, FsNet, Supervised CAE using the authors' best results. All three experiments follow the standard workflow:

\begin{itemize}
    \item Split each data sets into training and test partition.
    \item Run SABCE on the training set to extract top $K\in\{10,50\}$ features.
    \item Using the top $K$ features train a one hidden layer ANN classifier with $H$ ReLU units to predict the test samples. The $H$ is picked using a validation set.
    \item Repeat the classification 20 times and report average accuracy.
\end{itemize}
Now we describe the details of the two experiments.

\textbf{Experiment 1:} The first bench-marking experiment is conducted on six publicly available \citep{li2018feature} high dimensional biological data sets: ALLAML, GLIOMA, SMK\_CAN, Prostate\_GE, GLI\_85, and CLL\_SUB\footnote{Available at https://jundongl.github.io/scikit-feature/datasets.html} to compare SABCE with FsNet, Supervised CAE (SCAE), and Stochastic Gates (STG). Following the experimental protocol of Singh et al. \citep{singh2020fsnet}, we randomly partitioned each data into a 50:50 ratio of train and test and ran SABCE, STG on the training set. After that, we calculated the test accuracy using the top $K=\{10,50\}$ features. We repeated the experiment 20 times and reported the mean accuracy. We ran a 5-fold cross-validation on the training set to tune the hyperparameters.

\textbf{Experiment 2:} In the second bench-marking experiment, we compared our approach with LassoNet\citep{lemhadri2021lassonet} and Stochastic Gate\citep{yamada2020feature} on six data sets: Mice Protein\footnote{There are some missing entries that are imputed by mean feature values.}, COIL20, Isolet, Human Activity, MNIST, and FMNIST\footnote{Available at UCI Machine Learning repository}. Following the experimental set of Lemhadri et al., we split each data set into 70:10:20 ratio of training, validation, and test sets. We ran SCE on the training set to pick the top $K=50$ features to predict the class labels of the sequester test set. We extensively used the validation set to tune the hyperparameters.

\subsection{Results}
Now we discuss the results of the benchmarking experiments. In Table~ \ref{table:exp1_results} we present the results of the first experiment where we compare SABCE, SCAE, STG, and FsNet on six high-dimensional biological data sets. Apart from the results using a subset (10 and 50) of features, we also provide the prediction using all the features. In most cases, feature selection helps improve classification performance. Generally, SABCE features perform better than SCAE and FsNet; out of the twelve classification tasks, SABCE produces the best result on ten. Notice that the top fifty SABCE features give a better prediction rate than the top ten in all the cases. Interestingly, the accuracy of SCAE and FsNet drop significantly on SMK\_CAN, GLI\_85 and CLL\_SUB using the top fifty features.

\begin{figure}[ht!]
	\centering
	\includegraphics[width=12.5cm,height=9.5cm]{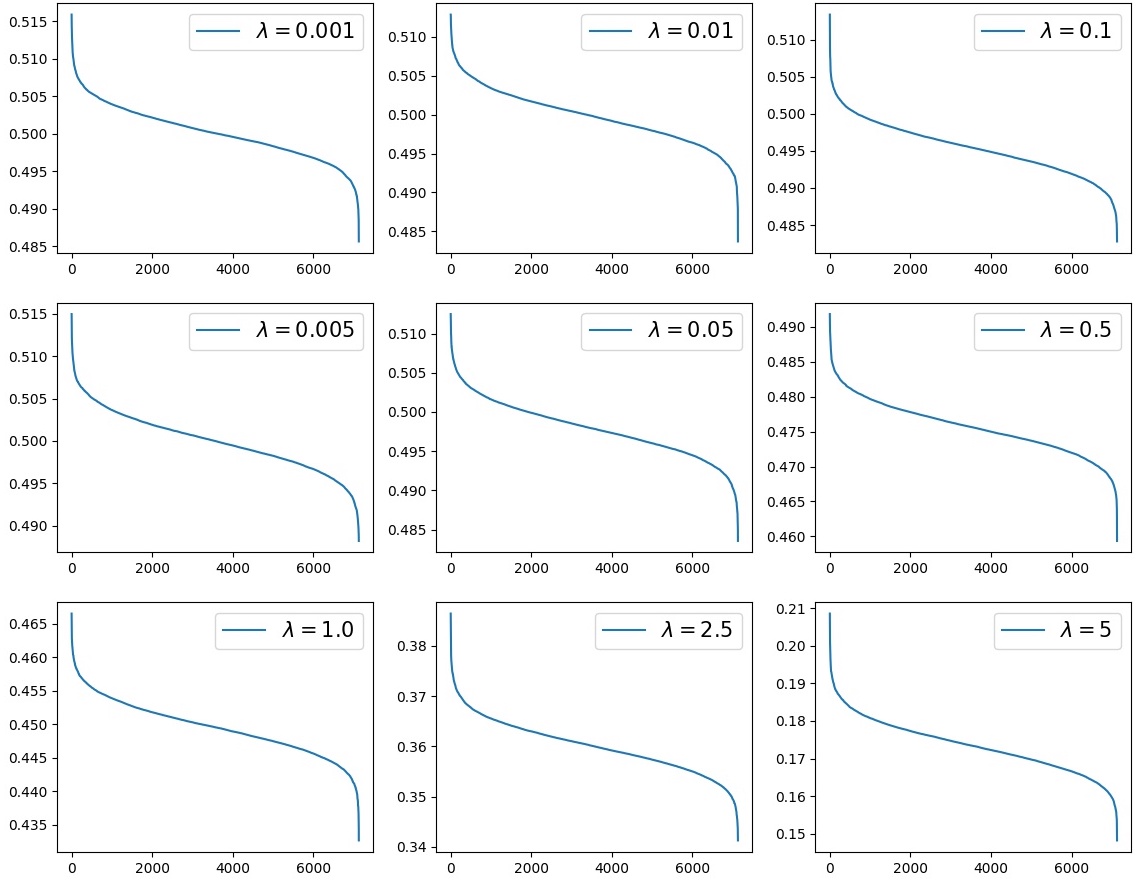}
	\caption{Sparsity analysis of Stochastic Gates on ALLAML data.}
	\label{fig:STG_Sparsity_ALLAML}
\end{figure}

\begin{table}[ht!]
	\vspace{-1.5mm}
	\centering
	\begin{tabular} {|c|c|c|c|c|c|c|c|c|c|}	
		\hline 	
		\multirow{2}{*}{Data set} & \multicolumn{4}{c|} {Top 10 features} & \multicolumn{4}{c|} {Top 50 features}  & \multirow{1}{*}{All} \\
		\cline{2-9}		
		& \multicolumn{1}{c|} {FsNet} &  \multicolumn{1}{c|} {SCAE} & \multicolumn{1}{c|} {STG} & \multicolumn{1}{c|} {SABCE} & \multicolumn{1}{c|} {FsNet} & \multicolumn{1}{c|} {SCAE} & \multicolumn{1}{c|} {STG} & \multicolumn{1}{c|} {SABCE} &  \multirow{1}{*}{Fea.}\\
		
		\hline
		ALLAML & $91.1$ & $83.3$ & $81.0$ & $\textbf{93.7}$ & $92.2$ & $93.6$ & $88.5$ & $\textbf{94.6}$ & $89.9$\\
        \hline
        Prostate\_GE & $87.1$ & $83.5$ & $82.3$ & $\textbf{89.9}$ & $87.8$ & $88.4$ & $85.0$ & $\textbf{90.1}$ & $75.9$\\
		\hline
        GLIOMA & $62.4$ & $58.4$ & $62.0$ & $\textbf{66.8}$ & $62.4$ & $60.4$ & $70.4$ & $\textbf{74.2}$ & $70.3$\\
		\hline
		SMK\_CAN & $\textbf{69.5}$ & $68.0$ & $65.2$ & $68.1$ & $64.1$ & $66.7$ & $68.0$ & $\textbf{69.4}$ & $65.7$\\
		\hline
        GLI\_85 & $87.4$ & $\textbf{88.4}$ & $72.2$ & $84.7$ & $79.5$ & $82.2$ & $81.0$  & $\textbf{85.7}$ & $79.5$\\
        \hline
        CLL\_SUB & $64.0$ & $57.5$ & $54.4$ & $\textbf{70.8}$ & $58.2$ & $55.6$ & $63.2$ & $\textbf{72.2}$ & $56.9$\\
        \hline
		
	\end{tabular}	
	\caption{Comparison of mean classification accuracy of FsNet, SCAE,STG and SABCE features on six real-world high-dimensional biological data sets. The prediction rates are averaged over twenty runs on the test set. Numbers for FsNet and SCAE are being reported from \citep{singh2020fsnet}. The last column reports accuracy using all features using an ANN classifier.}
	\label{table:exp1_results}
\end{table}
Now we turn our attention to the results of the second experiment, as shown in Table~\ref{table:exp2_results}. The features of the SABCE produce better classification accuracy than LassoNet in all cases. Besides COIL20, our model has better accuracy by a margin of $4\% - 6.5\%$. On the other hand, STG performed slightly better (a margin of 0.4\% to 0.7\%) than SABCE on Mice Protein, COIL20, and MNIST. In contrast, our model is more accurate than STG on FMNIST, ISOLET, and Activity by 1.6\% to 4.2\%. Note that LassoNet is the worst-performing model. In this experiment, STG performed competitively compared to the first, where STG's performance was significantly worse than that of SABCE. Upon further investigation, it turns out that the model fails to induce feature sparsity on all six high-dimensional biological data sets. We fit the model on the training partition of each data set and then plot the probability of the stochastic gates in descending order, which we call the {\textit {sparsity plot}}. We run STG using a wide range on $\lambda$, which controls the sparsity of the model. In Figure \ref{fig:STG_Sparsity_ALLAML}, we show the result on ALLAML data. As we can see, the model doesn't create a sparse solution for input features for any of the nine values. Ideally, we should have observed the probability of many variables to near zero so that those features could be ignored. Changing the activation function, number of hidden nodes, or depth of the network doesn't produce a sparser solution. We kept the similar analysis on other data sets in Appendix (Section \ref{STG_Sparsity}).

\begin{table}[ht!]
	\vspace{-3mm}
	\centering
	\begin{tabular} {|c|c|c|c|c|}	
		\hline
		\multirow{2}{*}{Data set} & \multicolumn{3}{c|} {Top 50 features} & \multirow{1}{*}{All features} \\
		\cline{2-4}
		& \multicolumn{1}{c|} {LassoNet} & \multicolumn{1}{c|} {STG} &  \multicolumn{1}{c|} {SABCE} & {ANN} \\
		
		\hline
		Mice Protein &$95.8$ & \textbf{99.8} & $99.4$ & $99.00$\\
		\hline
		MNIST & $87.3$ & $\textbf{94.7}$ & $94.0$ & $92.8$\\
		\hline
        FMNIST & $80.0$ & $83.8$ & $\textbf{85.4}$ & $88.30$\\
        \hline
        ISOLET & $88.5$ & $90.9$ & $\textbf{93.3}$ & $95.30$\\
        \hline
        COIL-20 & $99.1$ & $\textbf{99.7}$ & $99.3$ & $99.60$\\
        \hline
        Activity & $84.9$ & $86.6$ & $\textbf{90.8}$ &  $85.30$\\
		\hline
	\end{tabular}	
	\caption{Classification results using LassoNet, STG, and SABCE features on six publicly available data sets. Numbers for LassoNet and 'All features ANN' are reported from \citep{lemhadri2021lassonet}. All the reported accuracies are measured on the test set.}
	\label{table:exp2_results}
\end{table}

\section{Related Work}
\label{lit}

Feature selection has a long history spread across many fields, including bioinformatics, document classification, data mining, hyperspectral band selection, computer vision, etc. 
We describe the literature related to the embedded methods where the selection criteria are part of a model. The model can be either linear or non-linear. Adding an $\ell_1$ penalty to classification and regression methods naturally produce feature selectors for linear model, see \citep{tibshirani1996regression,fonti2017feature,muthukrishnan2016lasso,kim2004gradient,zou2005regularization,marafino2015efficient,shen2011identifying,sokolov2016pathway,lindenbaum2021randomly,candes2008enhancing,daubechies2010iteratively,bertsimas2017trimmed,xie2009scad}. Support Vector Machines \citep{cortes1995support} have been used extensively for feature selection, see
\citep{marafino2015efficient,shen2011identifying,sokolov2016pathway,guyon2002gene,o2013iterative,chepushtanova2014band}.

While the linear models are generally fast and convex, they don't capture the non-linear relationship among the input features (unless a kernel trick is applied). 
Non-linear models based on deep neural networks overcome these limitations. Here, we will briefly discuss a handful of such models. Group Sparse ANN  \citep{scardapane2017group} used group Lasso \citep{tibshirani1996regression} to impose the sparsity on a group of variables instead of a single variable. 
Li et al. proposed deep feature selection (DFS), which is a multilayer neural network-based feature selection technique \citep{li2016deep}. 
\citep{kim2016opening} proposed a heuristics based technique to assign importance to each feature. Using the ReLU activation, \citep{roy2015feature} provided a way to measure the contribution of an input feature towards hidden activation of next layer. \citep{han2018autoencoder} developed an unsupervised feature selection technique based on the autoencoder architecture. 
\citep{taherkhani2018deep} proposed a RBM \citep{Hinton:2006:FLA:1161603.1161605,hinton2006reducing} based feature selection model. Also see, \citep{balin2019concrete,yamada2020feature, singh2020fsnet}.

\section{Discussion, Conclusion and Limitations}
\label{dis_cons}
In this paper, we proposed a novel neural network-based feature selection technique, Sparse Adaptive Bottleneck Centroid-Encoder (SABCE). Using the basic multi-layer perceptron encoder-decoder neural network architecture, the model backpropagates the SABCE cost to a feature selection layer that filters out non-discriminating features by $\ell_{2,1}$-regularization. The setting allows the feature selection to be data-driven without needing prior knowledge, such as the number of features to be selected and the underlying distribution of the input features.
The extensive analysis in Section \ref{empirical_analysis} demonstrates that the $\ell_{2,1}$-norm induces good feature sparsity without shrinking all the variables. Unlike other methods, e.g., Stochastic Gates, our approach promotes feature sparsity for high-dimension and low-sample size biological datasets, further demonstrating the value of SABCE as a feature detector. We chose the $\lambda_1$ and $\lambda_2$ from the validation set from a wide range of values and saw that smaller values work better for classification. The plots with the Venn diagrams confirm the consistent and stable feature detection ability of SABCE.

The rigorous benchmarking with twelve data sets from diverse domains and four methods provides evidence that the features of SABCE produce better generalization performance than other state-of-the-art models. We compared SABCE with FsNet, mainly designed for high-dimensional biological data, and found that our proposed method outperformed it in most cases. In fact, our model produced new state-of-the-art results in ten cases out of twelve. The comparison also includes Supervised CAE, which is less accurate than SABCE. On the data sets where the number of observations is more than the number of variables, SABCE features produces better classification results than LassoNet in all six cases and better than Stochastic Gates on three data sets. The strong generalization performance, coupled with the ability to sparsify input features, establishes the value of our model as a nonlinear feature detector.

Although SABCE produces new state-of-the-art results on diverse data sets, our model won't be the right choice for the cases where class centroids make little sense, e.g., natural images. The current scope of the work doesn't allow us to investigate other optimization techniques, e.g., proximal gradient descent, trimmed Lasso, etc., which we plan to explore in the future.


\bibliography{SABCE}


\newpage
\appendix
\section{Reproducibility}
\label{reproducibility}
Table \ref{table:SCEHyperparameters_dataset} shows the values of the hyperparameters that we use in our experiments. The tuning of these parameters is done in two ways: for low sample size biological data sets (first six data sets in the Table 
\ref{table:SCEHyperparameters_dataset}), we run five-fold cross-validation on a training partition. We split each of the remaining data sets into train, validation, and test ratio of 70:10:20 and used the validation set for tuning.

\begin{table}[!ht]
	\centering
	\begin{tabular} {|c|c|c|c|c|c|c|}	
		\hline	
		Dataset & Network topology & Activation & Learning Rate & $\lambda_1,\lambda_2$ & $\mu_1,\mu_2$ & Epoch\\
		\hline
		ALLAML & $d \rightarrow 250 \rightarrow d$ & tanh & 0.001 & 0.001,0.001 & 0.6,0.1 & 1050\\
		GLIOMA & $d \rightarrow 250 \rightarrow d$ & tanh & 0.001 & 0.001,0.001 & 0.6,0.1 & 1050\\
		SMK\_CAN & $d \rightarrow 250 \rightarrow d$ & tanh & 0.001 & 0.001,0.001 & 0.6,0.1 & 1050\\
		Prostate\_GE & $d \rightarrow 250 \rightarrow d$ & tanh & 0.001 & 0.001,0.001 & 0.6,0.1 & 1050\\
		GLI\_85 & $d \rightarrow 250\rightarrow d$ & tanh & 0.001 & 0.001,0.001 & 0.6,0.1 & 1050\\
		CLL\_SUB & $d \rightarrow 250\rightarrow d$ & tanh & 0.001 & 0.001,0.001 & 0.6,0.1 & 1050\\
  \hline
        Mice Protein & $d \rightarrow 100 \rightarrow d$ & tanh & 0.008 & 0.001,0.001 & 0.2,0.6 & 1050\\
		COIL20 & $d \rightarrow 100 \rightarrow d$ & tanh & 0.008 & 0.001,0.001 & 0.8,0.1 & 1050\\
		Isolet & $d \rightarrow 100 \rightarrow d$ & tanh & 0.008 & 0.001,0.001 & 0.8,0.3 & 1050\\
		Human Activity & $d \rightarrow 100 \rightarrow d$ & tanh & 0.008 & 0.001,0.001 & 1.0,0.9 & 1050\\
		MNIST & $d \rightarrow 100 \rightarrow d$ & tanh & 0.008 & 0.001,0.001 & 0.6,0.1 & 1050\\
		FMNIST & $d \rightarrow 100 \rightarrow d$ & tanh & 0.008 & 0.001,0.001 & 0.6,0.1 & 1050\\
		\hline
	\end{tabular}	
 \medskip
	\caption{Details of network topology and hyperparameters for SABCE. The number $d$ is the input dimension of the network and is data set dependent.}
	\label{table:SCEHyperparameters_dataset}
\end{table}
\newpage
\section{Compute Time}
We present the average compute time of training for each data set in the table below.

\begin{table}[!ht]	
	
	\centering
	\begin{tabular} {|c|c|}	
		\hline	
		Dataset & Training time (in minutes)\\
		\hline
        ALLAML & 0.137 \\
        GLIOMA & 0.122 \\
        SMK\_CAN & 0.889 \\
        Prostate\_GE & 0.121 \\
        GLI\_85 &  0.809\\
        CLL\_SUB &  0.276\\
        Mice Protein &  1.538\\
        COIL20 &  2.563\\
        Isolet &  4.757 \\
        Human Activity &  2.795\\
        MNIST &  20.621\\
        FMNIST & 23.697 \\
		\hline
	\end{tabular}	
 \medskip
	\caption{Average training time for Sparse Adaptive Bottleneck Centroid-Encoder for each data set.}
	\label{table:SABCE_Computetime}
\end{table}
\newpage
\section{Sparsity Analysis of Stochastic Gates}
\label{STG_Sparsity}
In this section, we present a detailed analysis of the input feature sparsity of Stochastic Gates (STG)\citep{yamada2020feature}. We have observed that STG performed competitively for the data sets with more samples than features (Table 5 of main paper); in contrast, the model performed relatively weakly on data sets where the number of variables is very large than the number of input data (Table 4 of main paper). Upon further investigation, it turns out that the model fails to induce feature sparsity on all six high-dimensional biological data sets. We fit the model on the training partition of each data set and then plot the probability of the stochastic gates in descending order, which we call the {\textit {sparsity plot}}. We run STG using a wide range on $\lambda$, which controls the sparsity of the model. 

\begin{figure}[ht!]
	\centering
	\includegraphics[width=12.5cm,height=10.0cm]{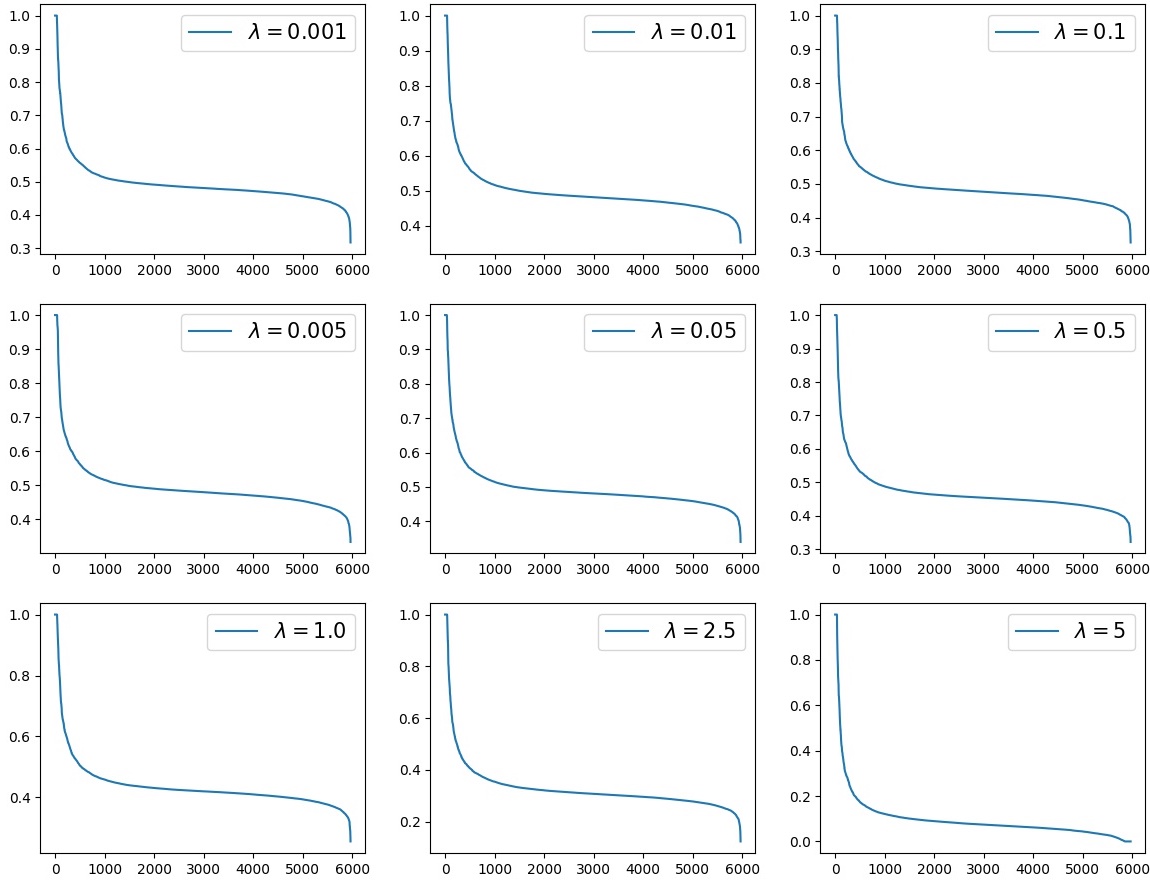}
	\caption{Sparsity analysis of Prostate\_GE data.}
	\label{fig:STG_Sparsity_Prostate_GE}
\end{figure}

\begin{figure}[ht!]
	\centering
	\includegraphics[width=12.5cm,height=9.5cm]{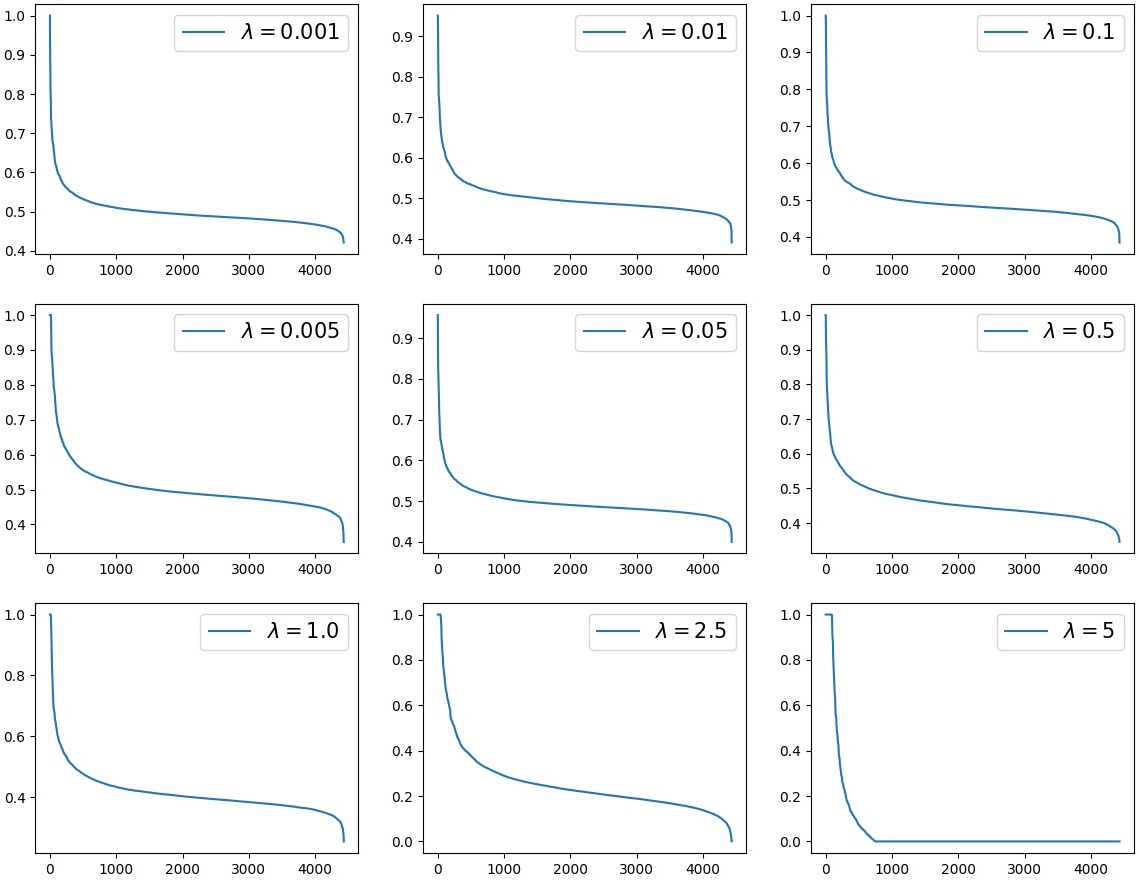}
	\caption{Sparsity analysis of GLIOMA data.}
	\label{fig:STG_Sparsity_GLIOMA}
\end{figure}

\begin{figure}[ht!]
	\centering
	\includegraphics[width=12.5cm,height=9.5cm]{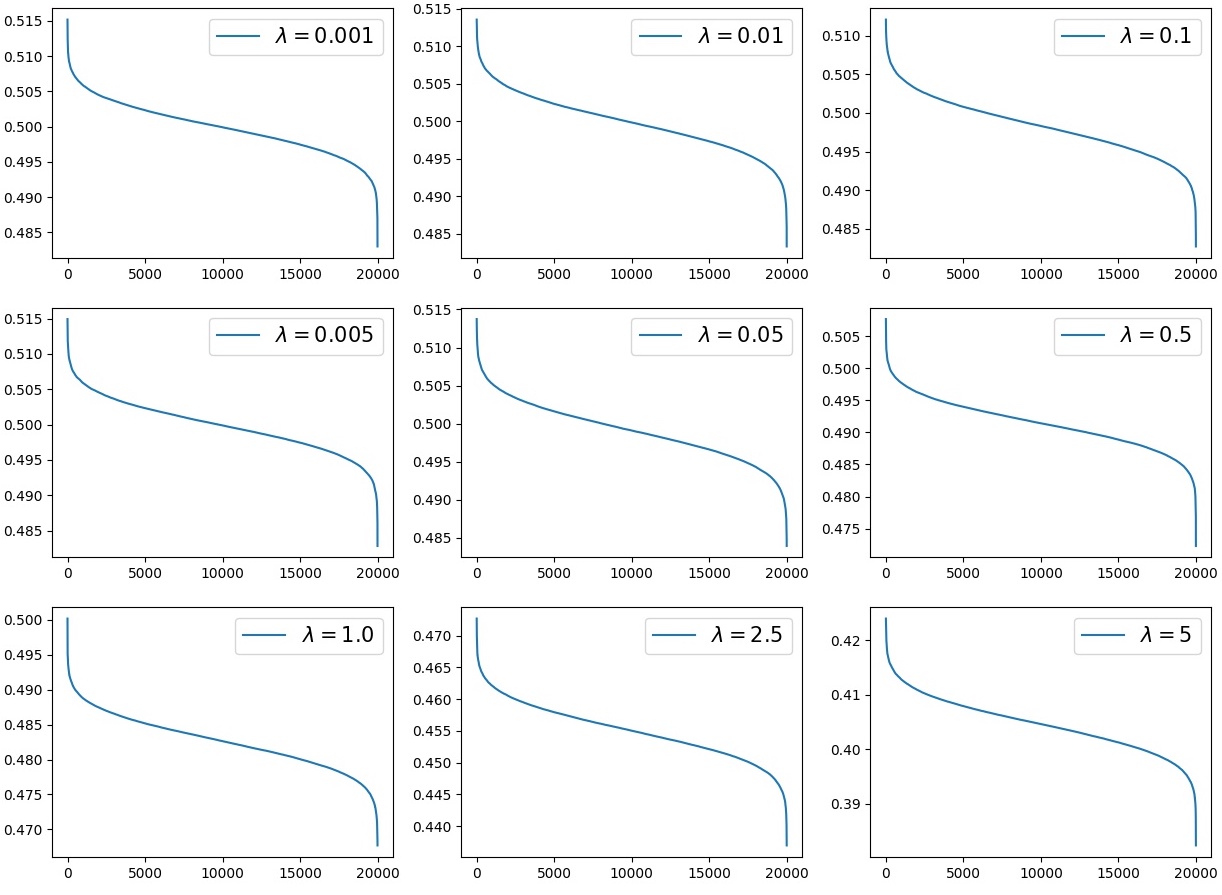}
	\caption{Sparsity analysis of SMK\_CAN data.}
	\label{fig:STG_Sparsity_SMK_CAN}
\end{figure}

\begin{figure}[ht!]
	\centering
	\includegraphics[width=12.5cm,height=9.5cm]{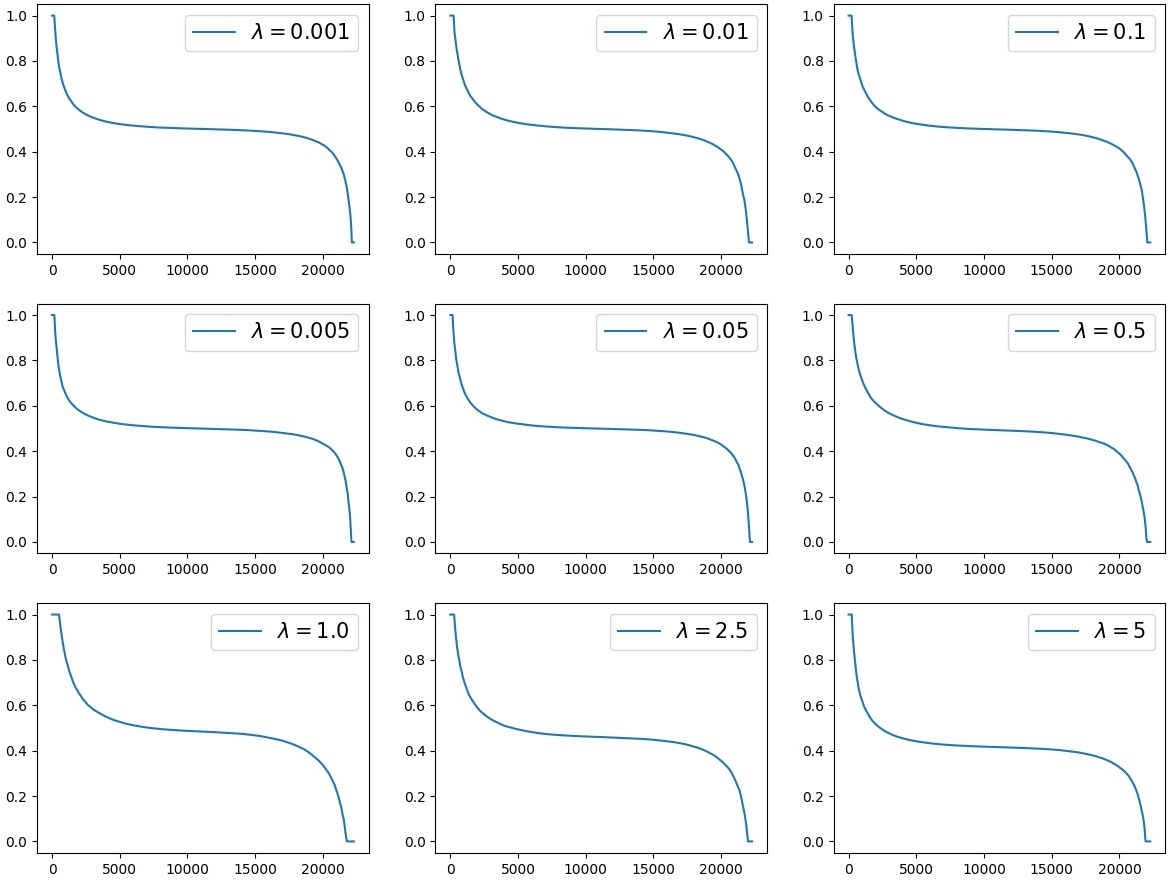}
	\caption{Sparsity analysis of GLI\_85 data.}
	\label{fig:STG_Sparsity_GLI_85}
\end{figure}

\begin{figure}[ht!]
	\centering
	\includegraphics[width=12.5cm,height=9.5cm]{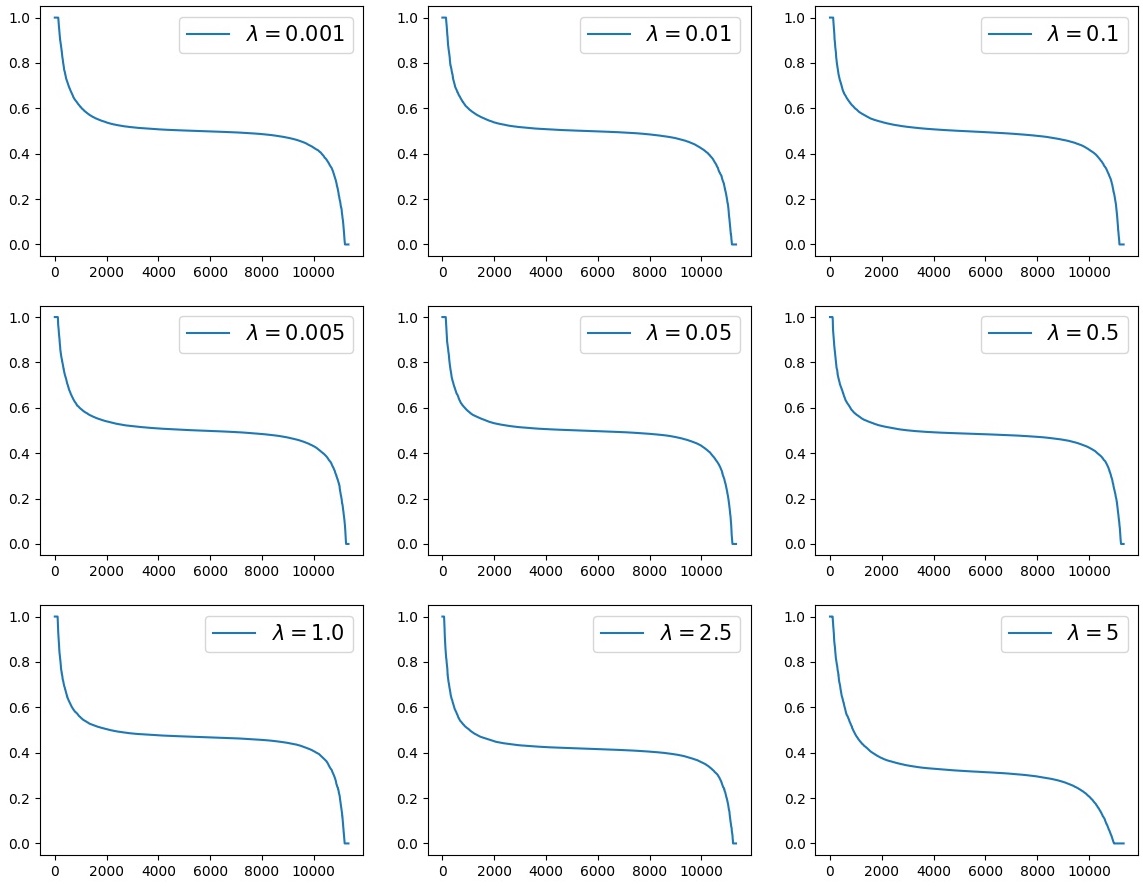}
	\caption{Sparsity analysis of CLL\_SUB data.}
	\label{fig:STG_Sparsity_CLL_SUB}
\end{figure}

In Figure \ref{fig:STG_Sparsity_Prostate_GE}, \ref{fig:STG_Sparsity_GLIOMA},\ref{fig:STG_Sparsity_SMK_CAN}, \ref{fig:STG_Sparsity_GLI_85} and \ref{fig:STG_Sparsity_CLL_SUB}, we present the sparsity plot for Prostate\_GE, GLIOMA, SMK\_CAN, GLI\_85, and CLL\_SUB data respectively. These sparsity plots are qualitatively similar to ALLAML (Figure \ref{fig:STG_Sparsity_ALLAML}), except for GLIOMA when $\lambda=5.0$.\\
It is clear from this analysis that Stochastic Gate cannot sparsify the hi-dimensional biological data sets, and it's plausible that the model doesn't always pick the proper discriminatory feature set. The classification results in Table 4 (in the main article) do support the claim.

\newpage

\end{document}